\documentclass[letterpaper, 10 pt, conference]{ieeeconf}  % Comment this line out if you need a4paper

\usepackage[usenames,dvipsnames]{color} %using the color package, not xcolor

\IEEEoverridecommandlockouts                              % This command is only needed if 
\usepackage{graphicx}

\title{\LARGE \bf
Distribution of Responsibility During the Usage of AI-Based Exoskeletons for Upper Limb Rehabilitation}

\author{Huaxi (Yulin) Zhang $^{1}$, Mélanie Fontaine $^{2}$,  Marianne Huchard $^{3}$, Baptiste Mereaux $^{4}$, Olivier Rémy-Neris $^{5}$ 
\thanks{$^{1}$ Huaxi (Yulin) Zhang is with Laboratory Of Innovative Technologies, University of Picardie Jules Verne, Saint Quentin, France.
        {\tt\small yulin.zhang@u-picardie.fr}}
\thanks{$^{2}$ Mélanie Fontaine is with Laboratory Of Innovative Technologies,
        University of Picardie Jules Verne, Saint Quentin, France.
        {\tt\small melanie.fontaine@etud.u-picardie.fr}} 
\thanks{$^{3}$ Marianne Huchard is with LIRMM, Université de Montpellier, CNRS, Montpellier, France
        {\tt\small marianne.huchard@lirmm.fr}}
\thanks{$^{4}$ Baptiste Mereaux is with Laboratory Of Innovative Technologies,
        University of Picardie Jules Verne, Saint Quentin, France.
        {\tt\small baptiste.mereaux@crispi-upjv.fr}}
\thanks{$^{5}$Olivier Rémy-Neris is with CHRU de Brest, Hôpital Morvan, Service de Médecine Physique et de Réadaptation, Brest, France {\tt\small olivier.remy-neris@univ-brest.fr}} % <-this % stops a space
}

\begin{document}

\maketitle
\thispagestyle{empty}
\pagestyle{empty}

%%%%%%%%%%%%%%%%%%%%%%%%%%%%%%%%%%%%%%%%%%%%%%%%%%%%%%%%%%%%%%%%%%%%%%%%%%%%%%%%
\begin{abstract}

%Les problèmes ethiques liées au developement d'un AI-base exoskeleton theorique et pratique ne sont pas %encore bien exprimé c'est pourquoi dans ce papier nous mettons en evidence les differents problème %ethique aussi bien sur le developement d'un AI-based exoskeleton que sur son utilisation. Ce papier %permet de donner un point de vue theorique concernant robotics, ai, healthcare, exoskeleton et %rehabilitation guidelines, mais aussi pratique avec l'utilisation de donnés aproprié grace au EMG ou %biomarkeur. L'objectif et de montrer que l'on peu avoir confiance en l'IA et que l'on reussir à la fois %avoir un AI-based exoskeleton fonctionnel et performant en prenant en compte les different problème %ethique lié.
%Beaucoup de questions sont soulevé dans ce papier et cherche à definir au mieux les probleme lié au %contact H/M, c'est pourquoi le projet AiBle et utilisé en case study et permet d'avoir un exemple concré %de AI-based exoskeleton pour discuté pas suelement de l'ethique mais aussi d'une implementation éthique.

The ethical issues concerning the AI-based exoskeletons used in healthcare have already been studied literally rather than technically. How the ethical guidelines can be integrated into the development process has not been widely studied. However, this is one of the most important topics which should be studied more in real-life applications. Therefore, in this paper we  highlight one ethical concern in the context of an exoskeleton used to train a user to perform a gesture: during the interaction between the exoskeleton, patient and therapist, how is the responsibility for decision making distributed? Based on the outcome of this, we will discuss how to integrate ethical guidelines into the development process of an AI-based exoskeleton. The discussion is based on a case study: AiBle. 
The different technical factors affecting the rehabilitation results and the human-machine interaction for AI-based exoskeletons are identified and discussed in this paper in order to better apply the ethical guidelines during the development of AI-based exoskeletons. 
%The objective of this paper is to discuss and propose a practical guidelines to develop human/machine interaction of the exos system while taking into account the ethical issue.

%Ethical, legal and societal implications (ELSI) in the development of wearable robots (WRs) are currently %not explicitly addressed in most guidelines for WR developers. Previous work has identified ELSI related %to WRs, e.g., impacts on body and identity, ableism, data protection, control and responsibilities, but %translation of these concerns into actionable recommendations remains outstanding. This paper provides %practical guidance for the implementation of ELSI in WR design, development and use. First, we identify %the need for domain-specific recommendations against the context of current ELSI guidance. We then %demonstrate the feasibility and usefulness of taking a domain-specific approach by successively %transforming currently identified ELSI into an action- guiding flowchart for integration of ELSI specific %to the different stages of WR development. This flowchart identifies specific questions to be considered %by WR development teams and suggests actions to be taken in response. By tailoring ELSI guidance to WR %developers, centring it on user needs, their relation to others and wider society, and being cognizant of %existing legislation and values, we hope to help the community develop better WRs that are safer, have %greater usability, and which impact positively on society~\cite{Kapeller2021}

\end{abstract}

%%%%%%%%%%%%%%%%%%%%%%%%%%%%%%%%%%%%%%%%%%%%%%%%%%%%%%%%%%%%%%%%%%%%%%%%%%%%%%%%
\section{INTRODUCTION}
%\textbf{more exoskeletons adopt, especially in healthcare domain.}
Exoskeletons are wearable systems, which can help human wearers perform a variety of tasks, such as carry heavy loads, reduce the burden of physically demanding tasks and also apply rehabilitation treatments to patients such as strokes~\cite{Rupal2017}.
The main thrust of research and development of exoskeletons has been focused on medical exoskeletons. 
Some medical exoskeletons are medical robots which are used to provide mobility to
physically disabled, injured or weakened people. 
% faut-il mettre que d'autres robots médicaux existent? par exemple les robots chirurgicaux mais que le plus souvent ce sont des manipulendum plus ou moins asservis?

The medical exoskeleton market size has increased very rapidly in recent years. It is projected to reach USD 1.0 billion by 2026, from USD 0.2 billion in 2021, at a CAGR of 45.0\%~\cite{Research2021}. The growth of this market has mainly been driven by an increase in the number of patients who can benefit of this technology and the subsequent growth in the demand for effective rehabilitation approaches. Agreements and collaborations between companies and research organizations for the development of the exoskeleton technology, and increasing insurance coverage for medical exoskeletons in several countries are the main factors of this increase. Most of the medical literature have observed a similar benefit of robotic and usual physiotherapy treatments.

Thanks to the development of artificial intelligence (AI) techniques, the adoption of AI in medical wearable exoskeletons has increased.
%As the development of AI is more and more mature, in the medical area, robotics combined to artificial intelligence are really hopefull.
The use of AI-based techniques has been recognised as enriching the rehabilitation process, providing a comprehensive assessment of a patient’s performance and increasing the confidence of end users i.e., patients and healthcare specialists, when interacting with robots for rehabilitation~\cite{VelezGuerrero2021}.

Meanwhile, the adoption of AI-based exoskeletons in rehabilitation poses real ethical concerns. 
These ethical issues have been studied and discussed in the literature recently~\cite{Longley2019,Kapeller2020,bentaieb2020augmented,Kapeller2021}. 
Kapeller \textit{et al.}~\cite{Kapeller2020} list 12 different ELS (Ethical, Legal and Social) issues, such as Body and Identity Impacts, the Experience of Vulnerability, Agency, Control and Responsibility, etc. The paper~\cite{Kapeller2021} presents the general ELSI (Ethical, Legal and Social Implications).
However, very little in-depth research has been done on how to implement these ethical guidelines technically. For example, in order to integrate/implement these ethical guidelines in robotics/AI development processes, which technical factors should be studied? 

Robotics systems with AI can be considered to be covering two overlapping sets of systems: systems that are only AI, systems that are only robotics. Thus the ethical issues concerning AI-based exoskeletons for rehabilitation is a very large subject. In this paper, we try to use a case study with a scenario to discuss one ethical issue as described in the following.

\begin{itemize}
    \item \textbf{Ethical questions posed by the project:} Responsibility distribution during the usage of exoskeletons. During rehabilitation exercise, if there are conflicts and different conclusions, who has the right to decide and how are responsibilities allocated among the different actors? 
    
    Who is the decision maker during the rehabilitation exercise, when there is a conflict? Patient? Therapist? Exoskeleton with AI?
    What role does AI play in this system? Is it a doctor, a patient, or a robot?
    \item \textbf{Technical viewpoint and factors: } From the Human-Machine interaction viewpoint, from the AI-based exoskeletons developer viewpoint, which factors will influence the rehabilitation results? These factors will decide how the system and AI model will be designed and developed. % factors.
    \item \textbf{Case study:} The AiBle~\footnote{This work was supported by EU project AiBle.}~\cite{AiBle} project aims to create an upper-limb exoskeleton using the latest technologies such as EMG (Electromyogram), AI, Cloud Computing and VR Games~\cite{AiBle}.

\end{itemize}

%structure of this paper
The paper is structured as follows. The AiBle case study and the questions posed by the project are presented in Section \ref{sec:casestudy}. Based on the case study, robotic and AI ethics guidelines and rules are discussed in Section \ref{sec:ethics}. In Section \ref{sec:discussion}, we will discuss which factors an AI-based  exoskeleton system can improve in rehabilitation. Section \ref{sec:technique} presents the technical factors which should be analyzed during the design of an AI-based exoskeleton system; and lastly, the conclusion which includes future work in Section \ref{sec:conclusion}.

\section{Case study: AiBle}
\label{sec:casestudy}

\begin{figure}[!thpb]
  \centering
  \includegraphics[width=0.98\linewidth]{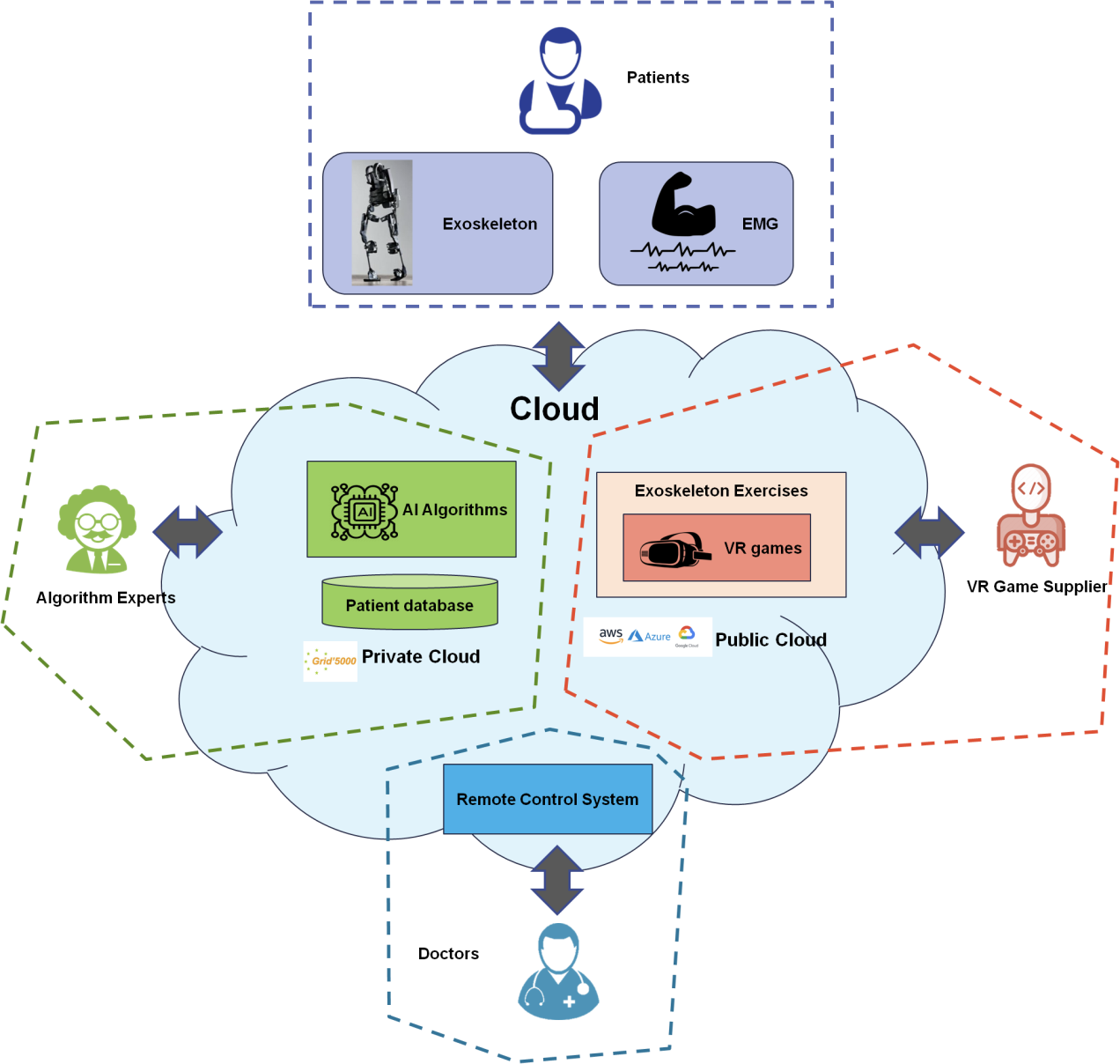}
  \caption{The overview of AiBle project with (AI, Cloud and VR)}
  \label{fig:aible}
\end{figure}

AiBle is a 3-year cross-border EU Interreg project between the UK and France. The aim of the project is to develop a new generation of robotic exoskeleton that benefits stroke patients by providing advanced functionality to enable remote but active rehabilitation.
This is achieved by the integration of artificial intelligence, virtual reality and cloud computing as shown in~\ref{fig:aible}. According to the difficulties of the ethical issues concerning Cloud robotics~\cite{FoschVillaronga2019}, in this paper, we only discuss the AI-based exoskeleton without the Cloud computing aspect.

\subsection{Question posed by the project}
\label{subsec:questions}

During the design of the human-machine interaction, the different system designers/developers and algorithm developers should answer the question:
\begin{itemize}
    \item \textit{Who will be the decision maker during the rehabilitation exercises, if there is a conflict? Patient? Therapist? Exoskeleton with AI?}
    \item \textit{What role does AI play in this system? Is it a doctor, a patient, a robot?}
\end{itemize}

% According to this question, we have launched a lot questions like.
% There is little empirical, social scientific data, for instance, about the responsibility distribution during the use of exos in rehabilitation: Can exoskeleton create a dependency? Can the doctor disagree with the use of an exoskeleton? Is the human contact an important factor for the success of the rehabilitation? Are the exoskeletons easier to use when the doctor was train on? If an issue occur can the patient trust the exoskeleton again? What kind of feed backs do we need to makes the AI efficient? What types of exercise the exoskeleton can totally replace? Whats the meaning of doing exercise in everyday life? Whats better between the diversity of exercise or the repetition with higher difficulties? Is the patient psychology the main factor? How many data do we need for the AI to reach the level of an expert? 

\subsection{AiBle decision making process}
Figure 2 explains the decision making process of the AiBle exoskeleton during the rehabilitation exercises. The decisions are about the difficulties of exercises and whether we need to change the exercises or the  exercise difficulty levels. In the project, we consider that the decision-making is carried out by the therapist.
\begin{enumerate}
    \item The AI-based exoskeleton gives feedback (sensor data and AI algorithm results) to the therapist. 
    \item The therapist discusses with the patient and takes their feelings/opinions into account.
    \item The therapist gives the diagnosis based on their experience and the feedback from the exoskeleton and the patient. 
    \item Finally, the therapist gives the final decision with the approval of the patient.
\end{enumerate} 

Thus, the AI-based exoskeleton is not 100\% autonomy level. The AI-based exoskeleton systems only give 1) the AI suggestions and 2) biomarkers collected by the exoskeleton to therapists. 
\begin{figure}[thpb]
  \centering
  \includegraphics[width=0.9\linewidth]{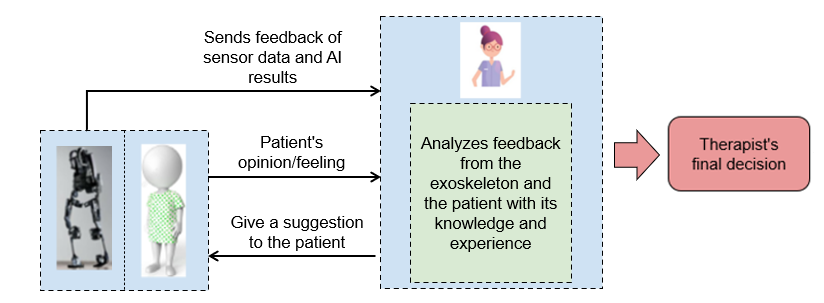}
  \caption{Decision making scenario of rehabilitation exercise in AiBle}
  \label{fig:scenario}
\end{figure}

However, although we have an idea about who will be the ultimate decision maker in our use case, the system design is still quite complicated.

\begin{itemize}
    \item What is the role of the AI in this scenario: is it a doctor, a patient, or a robot? In our system, AI should be trained using a dataset and then learns how to give correct diagnostics to the doctor and tries to simulate the doctor to make the best suggestions. Thus AI "considers" itself is a "doctor". This is not only a technical issue but also an ethical issue. 
    \item The AI-based exoskeleton is composed of the robot and the AI. Thus we cannot only consider robot ethics, but must also consider the ethics of AI.  
\end{itemize}

%However, although we have an idea about who will be the ultimate decision maker in our usecase,  how to design the AI-based exos help the therapist to take the best rehabilitation decisions which lead to the best rehabilitation results?

\section{Ethics Guidelines}
\label{sec:ethics}

An AI-based exoskeleton for rehabilitation is at the intersection of AI systems and robotics systems. Accordingly, the ethical issues of AI-based exoskeletons not only concern the ethical guidelines of robotics but also AI~\cite{Commission2018,Mueller2020,smuha2019eu, EthicsinScience2018}. In this paper, we concentrate on the distribution of the responsibility during the rehabilitation exercise and the ethical issues concerning AI-based exoskeleton design~\cite{Jobin2019,Floridi2022,MADIEGA2019}. 

\subsection{Robotics Guidelines}
In this paper, we use the original Asimov's laws~\cite{clarke1993asimov} instead of the newer one~\cite{5172885}, as in this paper the robot "considers" itself as a "doctor" and the final decision is made by person. 

\begin{itemize}
    \item A robot may not injure a human being or, through inaction, allow a human being to come to harm.
    \item A robot must obey the orders given it by human beings, except where such orders would conflict with the First Law.
    \item A robot must protect its own existence, as long as such protection does not conflict with the First or Second Laws. 
\end{itemize}

%There is ethical concerns about robotics:
%Firstly, rely on robots excessively may lead us to entrust it with sensitive tasks.
%Secondly, when failure occur, people will de-responsibilize themselves and put the blame on robots.
%Thirdly, robots could deskilling peoples even in critical context, but we need to keep them able to do %the jobs when robots can't.
%Finally, the way people will use it, robots owners can have bad intention- (The grand challenges of %Science Robotics)

\subsection{AI Guidelines}

It's important to define AI ethics especially when they are linked to a sensitive domain, such as in the case of healthcare exoskeletons. AI ethics have been widely studied in different works as discussed in~\cite{Jobin2019}. 
\cite{Floridi2022} has analyzed and summarized different AI ethical guidelines and proposed a framework based on five pillars: Beneficence, Non-maleficence, Autonomy, Justice, and Explicability.

\begin{itemize}
    \item \textbf{Beneficence vs Non-maleficence.} "Do only good" and "do not harm", are two different things. For example, "Do only good" is the criteria for the AI, if it finds that exercise A will be beneficial according to all the patients' rehabilitation histories. However, according to the patient's actual state, this may lead to harming the patient. Thus AI should choose from "do only good", and "do not harm".
    This is why we need to combine AI ethics with robotics ethics in this paper to analyze the AI-based robotic ethical problems. The robot must obey Asimov's the first law, the robot will never hurt a human, even if the AI results create conflicts, or if the order given by the therapist could be potentially harmful. We cannot assume the AI is designed perfectly and its prediction will "do only good" for patients. That's why the therapist's opinion and patient's opinion are important.  

    \item \textbf{Explicability.} A lot of work talks about the reasoning of AI decisions. The most important is, how do we interpret the AI result depending on which role the AI thinks it plays? Thus the question is "who is AI?". In our use case, is AI a doctor? a robot? or a patient? According to the different interpretations of AI's role in the AI-based robotic system, the system design will be completely different. The designers and developers of the AI-based robotics systems should consider this before they design the system. 
    
    \item \textbf{Autonomy.} After we know which role AI represents in the system, the next question is autonomy. Will we let the AI make the decision and control the robot 100\% or will there be a human intervention during the decision making or do we decide to adopt "decide-to-delegate"~\cite{Floridi2022}.

\end{itemize}

\subsection{Discussion of ethics concerning AI-based exoskeleton }
\label{subsec:discussion}
Among all the ethical issues of around AI-based robotics systems, we have identified three ethical issues which are related closely to the technical design: data privacy, evaluation of AI (trustworthiness of AI), and indirect AI manipulation.

\textbf{Data privacy.}
One more specific issue is that machine learning techniques in AI rely on training with vast amounts of data. This means there will often be a trade-off between privacy and rights to data vs. technical quality of the product. This influences the consequential evaluation of privacy-violating practices~\cite{Mueller2020}. 
However, PII (personally identifiable information) and HIPAA (Health Insurance Portability \& Accountability Act) are very important for AI-based exoskeletons to review and comply with. 

\textbf{Evaluation of AI.} Many AI systems rely on machine learning techniques in (simulated) neural networks that will extract patterns from a given dataset, with or without “correct” solutions provided,  i.e., supervised, semi-supervised or unsupervised. First of all, how to evaluate the correctness and effectiveness of the AI model?  Secondly, how to understand how it works is a difficult problem. Even a doctor sometimes makes a diagnosis with a unconscious bias but gives their reasons, so how can an AI composed of countless neural networks give you a reasonable explanation, this is an almost impossible proposition. 

\textbf{Indirect AI manipulation. } If therapists become accustomed to trusting the feedback results given by AI, will it cause therapists to become reliant on AI? This would allow AI to have more control over the entire rehabilitation process.
On the other hand, if the patient is too dependent on the robot, will it cause the patient to become addicted to robotic rehabilitation and thus affect their normal life, like the manipulations in games called “dark patterns”~\cite{Mathur2019}.

In the next sections, we will discuss the different factors which will influence the rehabilitation results using AI-based exoskeletons and how to apply ethical guidelines in the system design process.

\section{Different factors improved by adopting AI-based exoskeleton}
\label{sec:discussion}

Which factors affect clinical decision-making relating to the assessment of stroke rehabilitation? How to improve the rehabilitation results is key research in rehabilitation studies. %
Some work has been done on answering this question without the exoskeleton, such as~\cite{Longley2019}. Authors~\cite{Longley2019} have listed the different factors on three levels: patient-level (e.g. patient’s age), organizational (e.g. staffing levels) and clinician-level (e.g. experience). 

When using AI-based exoskeletons in rehabilitation exercises, we have identified three factors mentioned in the above work~\cite{Longley2019}, which could be improved.
\begin{itemize}
    \item \textbf{Patient motivation}. Patient motivation could be influenced by different factors, such as boredom. With AI-based exoskeletons, if the rehabilitation exercise is too simple, the system can give feedback to the therapist and allow the therapists to adapt the exercise level or change exercises. 
    \cite{Blanchard2022} illustrates that from the evaluation of the patients, one of the main advantages on the use of a robot is to propose diversified exercises and this influences the patient's adherence to the exercises.
    Such as, in our project by using VR games, we aim to motivate the patients to be interested in doing the exercises. If the therapists are over occupied or not very engaged with the patients due to pressure of time, the advantage of an AI-based exoskeleton is that it will be constantly involved with the patient to aid motivation.
    \item \textbf{Organizational service pressures}. Many organizations suffer from a shortage of therapists. With the AI-based exoskeleton, the therapists could be released from some of their stress and constant workload to allow their skills to be optimised.
    \item \textbf{Decision-making based on clinician’s knowledge/experience and their emotional state}. Firstly, the AI-based exoskeletons can return the biomarkers and AI analyzed results to the therapists, this information will help less experienced therapists to give a better diagnosis. Secondly, as mentioned in the previous item, the therapists will have less stressful working conditions with AI-based exoskeletons. 
\end{itemize}

%Thus, from a rehabilitation perspective, the effectiveness of a traditional treatment depends on the skill of therapists, their previous experience in treating similar cases, and their ability to formulate successful rehabilitation plans. Furthermore, the assessment of the patients and their progress is not quantified in a timely, adequate, and objective manner, thereby reducing the possibility of knowing the impact of rehabilitation.

There have already been encouraging results in the rehabilitation of upper limbs using only the application of assessment and therapeutic systems based on exoskeletons~\cite{Dellon2007,Pons2010,Ona2019}. We think that the AI-based exoskeleton will vastly improve the rehabilitation and treatment of patients.

\section{Technical factors during the system design}
\label{sec:technique}
In this section, we focus on discussing the technical factors from three viewpoints based on three roles: AI-based exoskeleton, Doctor/therapist, and patient.

\subsection{AI-based exoskeleton related factors}

\begin{itemize}
    \item \textbf{Data structure}. Which kinds of data should be collected? Why will this data be useful for AI to choose the most suitable rehabilitation exercises and evaluate the rehabilitation results? 
    
    The data could be divided into different categories: the general information concerning the patients themselves, the illness history (the type and severity of the stroke and the co-morbidities), and the data collected by the sensors. The important ones are the different biomarkers which could predict motor recovery~\cite{Stinear2017}?
    Are there any medical studies or clinical research supporting the choice of this data? 
    \item \textbf{Dataset quality}. Is the quality of the dataset good enough to have the best results? Does the dataset contain  bias (For example, if the dataset contains more diagnosis results from doctors who are less experienced than those who are more experienced)? 
    \item \textbf{AI method}. Which approach should AI developers choose: supervised, unsupervised or semi-supervised? For example unsupervised learning can degrade the confident level of AI model. 
    \item \textbf{Feedback of AI}. Which kind of information should be returned by AI? How could they help therapists/doctors to have a better evaluation of rehabilitation? 

\end{itemize}

\subsection{Therapist/Doctor-related factors}

\begin{itemize}
    \item \textbf{Feedback of AI-based exoskeletons.} Which kind of information do the therapists\& doctors require to evaluate the rehabilitation exercises, to choose a new exercise or to change the difficulty level of exercises? Here, we talk not only about the results from AI but also the raw data collected by exoskeletons such as different biomarkers that the therapist or doctors require to give a diagnosis. 
    \item \textbf{Therapist's understanding and training level on the exoskeleton.} The better the therapist's understanding of the exoskeleton, the better the outcome for the patient. A well trained therapist would know when a patient will benefit from exoskeleton-assisted therapy and integrate exoskeleton-assisted therapy into the overall treatment process because there are things robots can do better.
\end{itemize}

\subsection{Patient-related factors}

\begin{itemize}
    \item \textbf{Boredom level of rehabilitation exercises.} The exercises can become boring after a few sessions~\cite{Blanchard2022,Kenah2022}. When should AI or therapists/doctors suggest changing the exercises or increasing the exercise level?
    \item \textbf{Dementia.}  It is very important to know if patients have preexisting dementia. As the exoskeleton should be adjusted according to the level of dementia. 
\end{itemize}

When designing an AI-based exoskeleton system, the different actors such as AI algorithm scientists, HMI developers, robotic developers and doctors etc., should answer the two questions posed in Section~\ref{subsec:questions}. Then, during the system design, these factors should be taken into consideration before making the design decisions.

The above mentioned factors impact the effectiveness of the rehabilitation with AI-based exoskeletons. We have only listed the factors that we considered to be important, however this list is not exhaustive. 
% , these factors concern the patient as well as the AI-based exoskeleton and the doctor, therefore it is necessary to determine these factors in order to improve the treatment.
% These factors are mainly psychological, the patient's boredom to do the exercises is an important issue because that's the main factors to determine his will to succeed.
% These factors are not only on the patient's side, but also on the AI-based exoskeleton, which is the quality of the training and the information received by the AI thanks to the EMS or the biomarker.
% Finally, the training of the doctor on the exoskeleton and the confidence he has in the results of the exoskeleton are also factors not to be neglected.

%On parle des facteurs qui vont influencer les resultats de la rehabilitation selon les trois points de %vue patient docteur IA

\section{Conclusion}
\label{sec:conclusion}

In this paper, we have focused on discussing the technical factors for these actors: AI-based exoskeletons, therapists/doctors and patients. Two viewpoints have been considered: the distribution of responsibility during the usage and the human-machine interaction design for the usage of the exoskeletons. This is an open discussion issue, we do not give the answers in this paper. We have only listed the technical issues that should be considered and analyzed during the system design, based on our case study: AiBle. 

For future work, we want to study how to design AI-based exoskeleton with Cloud computing based on different ethical guidelines, in more depth.

Although there are a lot of ethical issues that should be discussed and ethical guidelines that have to be defined and implemented, the AI-based exoskeletons will be a life changing technique for rehabilitation treatment.

%\addtolength{\textheight}{-12cm}   % This command 

%%%%%%%%%%%%%%%%%%%%%%%%%%%%%%%%%%%%%%%%%%%%%%%%%%%%%

\bibliographystyle{IEEEtran}
\bibliography{IEEEabrv,ref}

\begin{thebibliography}{10}
\providecommand{\url}[1]{#1}
\csname url@samestyle\endcsname
\providecommand{\newblock}{\relax}
\providecommand{\bibinfo}[2]{#2}
\providecommand{\BIBentrySTDinterwordspacing}{\spaceskip=0pt\relax}
\providecommand{\BIBentryALTinterwordstretchfactor}{4}
\providecommand{\BIBentryALTinterwordspacing}{\spaceskip=\fontdimen2\font plus
\BIBentryALTinterwordstretchfactor\fontdimen3\font minus
  \fontdimen4\font\relax}
\providecommand{\BIBforeignlanguage}[2]{{%
\expandafter\ifx\csname l@#1\endcsname\relax
\typeout{** WARNING: IEEEtran.bst: No hyphenation pattern has been}%
\typeout{** loaded for the language `#1'. Using the pattern for}%
\typeout{** the default language instead.}%
\else
\language=\csname l@#1\endcsname
\fi
#2}}
\providecommand{\BIBdecl}{\relax}
\BIBdecl

\bibitem{Rupal2017}
B.~S. Rupal, S.~Rafique, A.~Singla, E.~Singla, M.~Isaksson, and G.~S. Virk,
  ``Lower-limb exoskeletons: Research trends and regulatory guidelines in
  medical and non-medical applications,'' \emph{International Journal of
  Advanced Robotic Systems}, vol.~14, no.~6, 2017.

\bibitem{Research2021}
Z.~M. Research, ``Medical exoskeleton market by component (hardware (sensor,
  actuator, control system, power source), software), type (powered, passive),
  extremities (lower, upper and full body) \& mobility (mobile, stationary) -
  global forecasts to 2026,'' \emph{Markets and Markets}, 2021.

\bibitem{VelezGuerrero2021}
M.~A. V{\'{e}}lez{-}Guerrero, M.~C. Cuervo, and S.~Mazzoleni, ``Artificial
  intelligence-based wearable robotic exoskeletons for upper limb
  rehabilitation: {A} review,'' \emph{Sensors}, vol.~21, no.~6, p. 2146, 2021.

\bibitem{Longley2019}
V.~Longley, S.~Peters, C.~Swarbrick, and A.~Bowen, ``What factors affect
  clinical decision-making about access to stroke rehabilitation? a systematic
  review,'' \emph{Clinical Rehabilitation}, vol.~33, no.~2, pp. 304--316, 2019,
  pMID: 30370792.

\bibitem{Kapeller2020}
A.~Kapeller, H.~Felzmann, E.~Fosch{-}Villaronga, and A.~Hughes, ``A taxonomy of
  ethical, legal and social implications of wearable robots: An expert
  perspective,'' \emph{Sci. Eng. Ethics}, vol.~26, no.~6, pp. 3229--3247, 2020.

\bibitem{bentaieb2020augmented}
K.~Bentaieb, ``The augmented human and artificial intelligence: what ethic for
  the human of the future? the example of the exoskeleton,'' \emph{Hitotsubashi
  Journal of Law and Politics}, vol.~48, pp. 63--67, 2020.

\bibitem{Kapeller2021}
A.~Kapeller, H.~Felzmann, E.~Fosch-Villaronga, K.~Nizamis, and A.-M. Hughes,
  ``Implementing ethical, legal, and societal considerations in wearable robot
  design,'' \emph{Applied Sciences}, vol.~11, no.~15, 2021.

\bibitem{AiBle}
\emph{https://www.euaible.com}.

\bibitem{FoschVillaronga2019}
E.~Fosch{-}Villaronga and C.~Millard, ``Cloud robotics law and regulation:
  Challenges in the governance of complex and dynamic cyber-physical
  ecosystems,'' \emph{Robotics Auton. Syst.}, vol. 119, pp. 77--91, 2019.

\bibitem{Commission2018}
E.~Commission, D.-G. for Research, Innovation, E.~G. on~Ethics~in Science, and
  N.~Technologies, \emph{Statement on artificial intelligence, robotics and
  'autonomous' systems : Brussels, 9 March 2018}.\hskip 1em plus 0.5em minus
  0.4em\relax Publications Office, 2018.

\bibitem{Mueller2020}
V.~C. M\"{u}ller, ``Ethics of artificial intelligence and robotics,'' in
  \emph{Stanford Encyclopedia of Philosophy}, E.~Zalta, Ed.\hskip 1em plus
  0.5em minus 0.4em\relax Palo Alto, Cal.: CSLI, Stanford University, 2020, pp.
  1--70.

\bibitem{smuha2019eu}
N.~A. Smuha, ``The eu approach to ethics guidelines for trustworthy artificial
  intelligence,'' \emph{Computer Law Review International}, vol.~20, no.~4, pp.
  97--106, 2019.

\bibitem{EthicsinScience2018}
E.~G. on~Ethics~in Science, N.~Technologies \emph{et~al.}, \emph{Statement on
  artificial intelligence, robotics and'autonomous' systems: Brussels, 9 March
  2018.}\hskip 1em plus 0.5em minus 0.4em\relax EU: European Union, 2018.

\bibitem{Jobin2019}
A.~Jobin, M.~Ienca, and E.~Vayena, ``The global landscape of {AI} ethics
  guidelines,'' \emph{Nat. Mach. Intell.}, vol.~1, no.~9, pp. 389--399, 2019.

\bibitem{Floridi2022}
L.~Floridi and J.~Cowls, \emph{A Unified Framework of Five Principles for AI in
  Society}.\hskip 1em plus 0.5em minus 0.4em\relax John Wiley \& Sons, Ltd,
  2022, ch.~22, pp. 535--545.

\bibitem{MADIEGA2019}
T.~A. MADIEGA, ``Eu guidelines on ethics in artificial intelligence: Context
  and implementation,'' EPRS: European Parliamentary Research Service, 2019.

\bibitem{clarke1993asimov}
R.~Clarke, ``Asimov's laws of robotics: implications for information
  technology-part i,'' \emph{Computer}, vol.~26, no.~12, pp. 53--61, 1993.

\bibitem{5172885}
R.~Murphy and D.~D. Woods, ``Beyond asimov: The three laws of responsible
  robotics,'' \emph{IEEE Intelligent Systems}, vol.~24, no.~4, pp. 14--20,
  2009.

\bibitem{Mathur2019}
A.~Mathur, G.~Acar, M.~Friedman, E.~Lucherini, J.~R. Mayer, M.~Chetty, and
  A.~Narayanan, ``Dark patterns at scale: Findings from a crawl of 11k shopping
  websites,'' \emph{Proc. {ACM} Hum. Comput. Interact.}, vol.~3, no. {CSCW},
  pp. 81:1--81:32, 2019.

\bibitem{Blanchard2022}
A.~Blanchard, S.~M. Nguyen, M.~Devanne, M.~Simonnet, L.~Goff-Pronost,
  O.~R{\'e}my-N{\'e}ris \emph{et~al.}, ``Technical feasibility of supervision
  of stretching exercises by a humanoid robot coach for chronic low back pain:
  The r-cool randomized trial,'' \emph{BioMed research international}, vol.
  2022, 2022.

\bibitem{Dellon2007}
B.~Dellon and Y.~Matsuoka, ``Prosthetics, exoskeletons, and rehabilitation
  [grand challenges of robotics],'' \emph{{IEEE} Robotics Autom. Mag.},
  vol.~14, no.~1, pp. 30--34, 2007.

\bibitem{Pons2010}
J.~L. Pons, ``Rehabilitation exoskeletal robotics,'' \emph{IEEE Engineering in
  Medicine and Biology Magazine}, vol.~29, no.~3, pp. 57--63, 2010.

\bibitem{Ona2019}
E.~D. Oña, J.~M. Garcia-Haro, A.~Jardón, and C.~Balaguer, ``Robotics in
  health care: Perspectives of robot-aided interventions in clinical practice
  for rehabilitation of upper limbs,'' \emph{Applied Sciences}, vol.~9, no.~13,
  2019.

\bibitem{Stinear2017}
C.~M. Stinear, ``Prediction of motor recovery after stroke: advances in
  biomarkers,'' \emph{The Lancet Neurology}, vol.~16, no.~10, pp. 826--836,
  2017.

\bibitem{Kenah2022}
K.~Kenah, J.~Bernhardt, N.~J. Spratt, C.~Oldmeadow, and H.~Janssen,
  ``Depression and a lack of socialization are associated with high levels of
  boredom during stroke rehabilitation: An exploratory study using a new
  conceptual framework,'' \emph{Neuropsychological Rehabilitation}, pp. 1--31,
  2022.

\end{thebibliography}

\end{document}